\documentclass[conference]{IEEEtran}
\IEEEoverridecommandlockouts
\usepackage{cite}
\usepackage{amsmath,amssymb,amsfonts,amsthm}
\usepackage{bm}
\usepackage{algorithmic}
\usepackage{graphicx}
\usepackage{textcomp}
\usepackage{xcolor}
\usepackage{bbm}
\usepackage{dsfont}
\usepackage{subcaption}
\usepackage{url}
\usepackage{hyperref}
\usepackage{cleveref}
\newtheorem{lemma}{Lemma}
\usepackage[font=small]{caption}

\usepackage{booktabs,makecell,multirow,hhline,threeparttable} 
\usepackage{xcolor,colortbl,pgf}
\definecolor{Color1}{RGB}{240, 240, 240}
\usepackage{makecell}

\usepackage{hyperref}

\newcommand{\supp}{\ensuremath{\mathop{\mathrm{supp}}}}

\newcommand{\proj}{\mathop{\mathrm{\Pi}}}
\newcommand{\id}{\mathds{1}}
\newcommand{\reals}{\mathbb{R}}
\newcommand{\naturals}{\mathbb{N}}
\renewcommand{\paragraph}[1]{\noindent\textbf{#1}}

\DeclareMathOperator*{\argmin}{arg\,min}
\DeclareMathOperator{\Tr}{Tr}

\newcommand\blfootnote[1]{%
  \begingroup
  \renewcommand\thefootnote{}\footnote{#1}%
  \addtocounter{footnote}{-1}%
  \endgroup
}

\def\BibTeX{{\rm B\kern-.05em{\sc i\kern-.025em b}\kern-.08em
    T\kern-.1667em\lower.7ex\hbox{E}\kern-.125emX}}

\newcounter{packednmbr}

\DeclareUnicodeCharacter{2212}{-}

\begin{document}
\bstctlcite{IEEEexample:BSTcontrol}

\title{Learn-Prune-Share for Lifelong Learning 
}

\author{\IEEEauthorblockN{Zifeng Wang\textsuperscript{1,*}, Tong Jian\textsuperscript{1,*}, Kaushik Chowdhury\textsuperscript{1}, Yanzhi Wang\textsuperscript{2}, Jennifer Dy\textsuperscript{1}, Stratis Ioannidis\textsuperscript{1}}
\IEEEauthorblockA{\textit{Department of Electrical and Computer Engineering} \\
\textit{Northeastern University}\\
Boston, MA \\
\textsuperscript{1}\{zifengwang, jian, krc, jdy, ioannidis\}@ece.neu.edu}
\textsuperscript{2}yanz.wang@northeastern.edu 
\thanks{
\textsuperscript{*} Z.Wang and T.Jian contributed equally to the paper.}
}

\maketitle

\begin{abstract}
\blfootnote{Accepted to the IEEE International Conference on Data Mining 2020 (ICDM'20)} In lifelong learning, we wish to maintain and update a model (e.g., a neural network classifier) in the presence of new classification tasks that arrive sequentially. In this paper, we propose a \emph{learn-prune-share} (LPS) algorithm which addresses the challenges of catastrophic forgetting, parsimony, and knowledge reuse simultaneously. LPS splits the network into task-specific partitions via an ADMM-based pruning strategy. This leads to \emph{no forgetting}, while maintaining \emph{parsimony}. Moreover, LPS integrates a novel selective knowledge sharing scheme into this ADMM optimization framework. This enables \emph{adaptive knowledge sharing} in an end-to-end fashion. Comprehensive experimental results on two lifelong learning benchmark datasets and a challenging real world radio frequency fingerprinting dataset are provided to demonstrate the effectiveness of our approach. Our experiments show that LPS consistently outperforms multiple state-of-the-art competitors.
\end{abstract}


\section{Introduction}

Human beings have a natural ability to adapt to different tasks sequentially without forgetting what they have learned. They can also seamlessly leverage knowledge learned from past tasks to tackle new tasks. This impressive ability is crucial for  learning systems deployed in the real world. 
Lifelong learning \cite{thrun1995lifelong}   aims to develop models that mimic this human ability to learn continually without forgetting knowledge acquired earlier.
%
In concrete terms, in a lifelong learning setting, we wish to maintain and update a model (e.g., a neural network classifier) in the presence of new classification tasks that arise sequentially. The model should  both exhibit high accuracy on new tasks as well as perform well on old classification tasks, even if the old data is no longer accessible. However, learning algorithms are often designed to operate under stationary data distributions -- typically, only a single task needs to be addressed. Under the lifelong learning setting, applying standard learning algorithms may lead to \emph{forgetting} what has been learned on old tasks: this phenomenon, known as \emph{catastrophic forgetting} \cite{mccloskey1989catastrophic, ratcliff1990connectionist}, results in severe performance degradation on old tasks after adapting to a new task.  

A large body of work has been proposed to address catastrophic forgetting, using a varied arsenal of techniques  \cite{parisi2019continual}. 
Despite advances in lifelong learning, there are still limitations. Most of the methods, including, e.g., regularization-based \cite{kirkpatrick2017overcoming, zenke2017continual, li2017learning, nguyen2017variational, aljundi2018memory} and rehearsal-based \cite{shin2017continual, lopez2017gradient, van2018generative, ostapenko2018learning} methods,
mitigate catastrophic forgetting under relatively restictive conditions, e.g., assuming a small number of highly related tasks. When tasks differ drastically, and the number of tasks grows, these methods suffer significant degradation. Another approach is to increase the model capacity (i.e., add parameters, neurons, layers, etc.), to accommodate new tasks, while preserving parts of the model for old tasks \cite{rusu2016progressive, yoon2017lifelong,draelos2017neurogenesis}. However, increasing complexity makes such methods prone to overfitting, and can be undesirable when models are to be deployed over memory-limited devices. Therefore, a competing objective of \emph{parsimony} is 
desirable. 

Another related challenge in lifelong learning is how to \emph{reuse} learned knowledge to help the model learn future tasks better. Current research work often ignores this critical point by, e.g., independently considering different tasks \cite{CLNP}, or by addressing it only partialy, e.g., using past parameters as an initialization during training  \cite{mallya2018packnet}. 
However, the usefulness of knowledge gained from old tasks may depend on the relevance between old and new tasks. For example, a classifier trained for classifying dogs may be more helpful for classifying cats than digits. Thus, how to \emph{adaptively select} useful past knowledge  is critical for improving the performance  on a new task.

Our proposed method, named \emph{learn-prune-share} (LPS), is a novel deep learning framework aimed at addressing these challenges. LPS learns sequential tasks without experiencing catastrophic  forgetting, by partitioning the neural network and dedicating portions to each task. It also prunes the neural network,  thereby maintaining parsimony and avoiding overfitting. Finally, it selectively shares knowledge from old tasks and reuses them on new tasks. All of these happen simultaneously, in a unified optimization framework trained in an end-to-end fashion. Our contributions are as follows:
\begin{itemize}
    \item We incorporate the state-of-the-art Alternating Direction Method of Multipliers (ADMM) based pruning strategy to solve the lifelong learning problem, maintaining a single parsimonious neural network model and eliminating catastrophic forgetting thoroughly.
    \item We design a novel knowledge sharing scheme, which learns to select useful knowledge from old tasks and adapt them to the current task. Our knowledge-sharing scheme is seamlessly integrated with our ADMM pruning strategy, and is trained jointly with the classifier parameters. We make our code publicly available\footnote{\texttt{\url{https://github.com/neu-spiral/LPSforLifelong}}} to accelerate community contributions in this exciting topic.
    \item Our method, LPS, shows superior performance on two standard lifelong learning benchmark datasets as well as a challenging real world radio fingerprinting dataset. LPS beats   state-of-the-art methods by a  2\%--54\% margin. 
\end{itemize}



\section{Related Work}

\subsection{Lifelong Learning}
\emph{Regularization-based} methods \cite{kirkpatrick2017overcoming, zenke2017continual, li2017learning, nguyen2017variational, aljundi2018memory}  
limit plasticity of the network via regularization terms or by limiting the learning rate on parameters learned from previous tasks.
While regularization-based methods  
mitigate catastrophic forgetting to some extent, performance on previous tasks gets increasingly worse when more diverse tasks are seen. By design, our method deals with catastrophic forgetting problem more effectively, as performance on previous tasks remains unchanged.

\emph{Rehearsal-based} methods capture the data distribution in previous tasks by learning a generative model.  When a new task arrives, data from previous tasks is simulated via the generative model and combined with current data to reinforce previous knowledge \cite{shin2017continual, lopez2017gradient, van2018generative, ostapenko2018learning}. 
Though saving the generative model is less memory intensive than saving data, such models can still be big. Performance largely depends on the quality generative model on careful tuning of the mix of generated and new data. Our approach avoids the additional cost of training and storing an external generative model, again while experiencing no catastrophic forgetting.

\emph{Expansion-based} methods accommodate new tasks by gradually increasing capacity of the model \cite{rusu2016progressive, yoon2017lifelong,draelos2017neurogenesis}. These methods generally outperform regularization and rehearsal based methods, which maintain a model with fixed capacity. However,  the size of model parameters grows linearly with the number of tasks. This limits their practical usage, and makes them prone to overfitting. On the contrary, our approach fully exploits the potential of a fixed-capacity model.

Our method is closest to Continual Learning via Neural Pruning (CLNP) \cite{golkar2019continual} and PackNet \cite{mallya2018packnet}. In these works, model pruning techniques are utilized to compress the original model iteratively to allocate free capacity for new tasks. However, both of these methods use simple threshold-based heuristics to prune the model with no structure constraint, resulting in a sparse, irregular matrix which limits further acceleration at inference time. Additionally, both of these methods consider tasks independently, ignoring the relationship between the current and previous tasks. In contrast, our approach adopts a systematic pruning strategy via Alternating Direction Method of Multipliers (ADMM), where structural constraints, e.g. filter pruning or column pruning \cite{ye2018progressive-pruning}, can be specified as needed. Moreover, our proposed novel knowledge inheritance scheme  \emph{adaptively select} weights shared from previous tasks to facilitate learning the current and future tasks.  Our experimental results in \Cref{sec:experiments} show that, due to these improvements, LPS  outperforms these two algorithms. 

\subsection{Neural Network Weight Pruning} 

The rich literature in neural network weight pruning can be categorized into {\emph{heuristic pruning algorithms}} and {\emph{regularization-based pruning algorithms}}. The former starts from the early work on irregular, unstructured weight pruning where arbitrary weights can be pruned. Han et al.~\cite{han2015deep} use an iterative algorithm to eliminate weights with small magnitude and perform retraining to regain accuracy. Guo et al.~\cite{guo2016dynamic} incorporate connection splicing into the pruning process to dynamically recover the pruned connections that are found to be important. Later, heuristic pruning algorithms have been generalized to the more hardware-friendly structured sparsity schemes. A Transformable Architecture Search (TAS)~\cite{dong2019network} realizes the pruned network and knowledge is transferred from the unpruned network to the pruned version. Luo et al.~\cite{luo2017thinet} leverage a greedy algorithm to guide the pruning of the current layer with input information of the next layer, while Yu et al.~\cite{yu2018nisp} define a ``neuron importance score" and propagate this score to conduct the weight pruning process.

Regularization-based pruning algorithms, on the other hand, have the unique advantage for dealing with structured pruning problems through group Lasso regularization \cite{liu2018rethinking}. Early work \cite{wen2016learning,he2017channel} incorporate $\ell_1$ or $\ell_2$ regularization in loss function to solve filter/channel pruning problems. Zhuang et al.~\cite{zhuang2018discrimination} introduce an $\ell_2$-norm variant indicating the number of selected channels in each layer. 
A number of subsequent works are dedicated to making the regularization penalty a dynamic and ``soft" term. The method in~\cite{he2018soft} selects filters based on $\ell_2$-norm and updates the filters that have been previously pruned, while \cite{zhang2018systematic,li2019compressing} incorporate the advanced optimization solution framework Alternating Direction Methods of Multipliers (ADMM) to achieve dynamic regularization penalty, thereby improving accuracy. 
%
We take advantage of the state-of-the-art ADMM-based pruning strategy by  \cite{zhang2018systematic} and \cite{li2019compressing}. Moreover, we integrate a novel selective knowledge sharing scheme into the ADMM optimization framework, captured by learnable masks. Furthermore, our whole pipeline can be trained in an end-to-end fashion performing learn, prune, share simultaneously through ADMM.


\section{Problem Formulation}
\label{sec:problem_formulation}

\begin{figure}[!t]
    \centering
    \includegraphics[width=0.6\columnwidth]{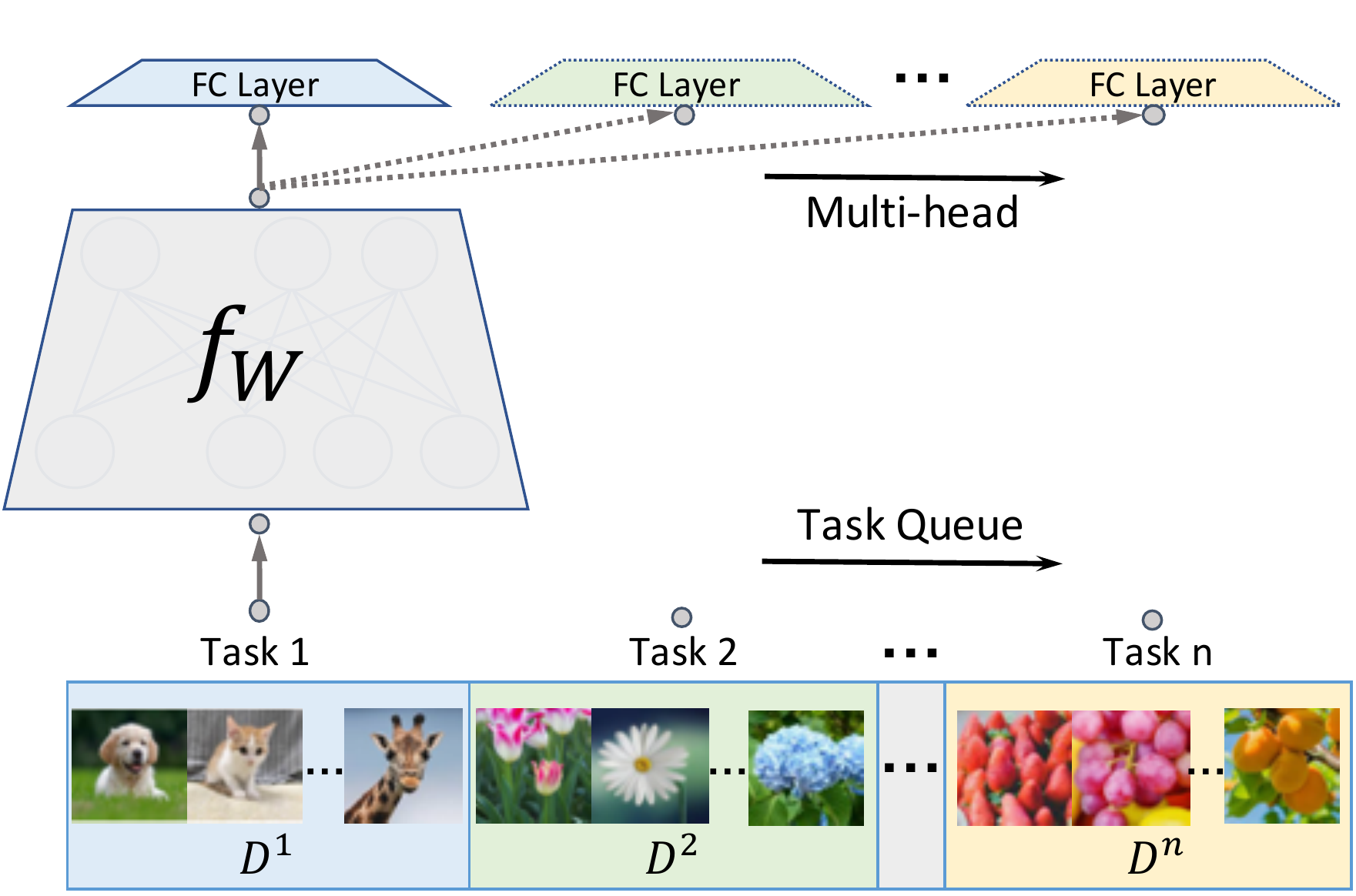}
    \caption{An illustration of supervised lifelong learning. A feature map $f_W$ is trained sequentially on datasets $\mathcal{D} = \{\mathcal{D}^1, \mathcal{D}^2,\ldots, \mathcal{D}^n\}$, where each dataset becomes accessible only at the corresponding task. A fully connected layer at the end of the classifier, denoted as one `head', is attached to $f_W$ to handle the new task. This is commonly referred to as a ``multi-head'' output later: faced with sequential $n$ tasks, the classifier branches in $n$ heads/output layers.}
    \label{fig:lifelong}
\end{figure}

In supervised lifelong learning, we are given a sequence of datasets $\mathcal{D} = \{\mathcal{D}^1, \mathcal{D}^2,\ldots, \mathcal{D}^n\}$, where each dataset $\mathcal{D}^t = \{(\mathbf{x}_i, y_i)\}_{i=1}^{m_t}$, 
$t=1,\ldots,n$, 
contains tuples of the input feature $\mathbf{x} \in \mathbb{R}^d$ and its corresponding label $y \in \mathbb{N}$. Each dataset corresponds to a distinct classification task: labels $y\in \mathbb{N}$ are disjoint across datasets $\mathcal{D}^t$. 
Datasets are revealed \textit{sequentially}: dataset $\mathcal{D}^t$ becomes  accessible only at the $t$-th task, which corresponds to, e.g., moving to a new environment. Our goal is to train a classifier sequentially on the datasets such that it achieves good performance on all tasks. 

Formally, we are given a feature extractor $f_W : \mathbb{R}^d \rightarrow \mathbb{R}^{d'}$ parameterized by $W\in \reals^m$.
After the network is trained on $\mathcal{D}^t$, along with a task-specific output layer,  its parameters $W$ are updated. If $W^t$ are the parameters of the feature extractor at task $t$, a final classifier is obtained after training the extractor (and the $n$ correponding output layers) on all datasets in $\mathcal{D}$ sequentially, as illustrated in Fig.~\ref{fig:lifelong}. The overall performance of $f_{W^{n}}$ is then assessed via the average  classification accuracy on separate testsets, one for each task $t$. Note that, at test time, we are aware of which task/environment $f_{W^n}$ is operating over, so that we can classify using the appropriate output layer.

While the problem setting is straightforward, we need to point out three desiderata that must be addressed by a supervised lifelong learning solution.

\noindent\textbf{Catastrophic Forgetting.} Catastrophic forgetting is the widely reported phenomenon \cite{mccloskey1989catastrophic,ratcliff1990connectionist} that  models, especially neural networks, tend to ``forget'' information from previous tasks when incorporating knowledge from new tasks. This is observed in accuracy performance degradation on previous tasks after being exposed to new tasks. Addressing catastrophic forgetting is a central issue, and the main objective of most lifelong learning algorithms \cite{rusu2016progressive, yoon2017lifelong,draelos2017neurogenesis,golkar2019continual,mallya2018packnet}.

\noindent\textbf{Parsimony.}  Due to limited computation and memory in real world applications, but also to avoid overfitting, the model $f_{{W}}$ should be as compact as possible. It is therefore desirable to maintain a single model and adapt it to various tasks, instead of, e.g., training multiple specialized models.

\noindent\textbf{Knowledge Reuse.} 
Related to both parsimony and catastrophic forgetting, beyond memorizing  knowledge acquired from previous tasks, we also want to exploit it when encountering new tasks. For example,  parts of the model could be shared across tasks; this  leverages relevant/reusable features across tasks, leading to further parsimony and avoiding overfitting, while also ameliorating catastrophic forgetting. Thus, it is important to strike a balance between  reuse vs.~growth or plasticity in a network, in a way that performance improves.

\section{Learn-Prune-Share}

We propose a learn-prune-share (LPS) algorithm, a novel deep learning framework for lifelong learning incorporating neural network pruning via ADMM. Our method maintains a single neural network for the sequence of tasks, while \textbf{learning} the tasks, \textbf{pruning} the neural network, and \textbf{sharing} knowledge among tasks; these three happen synergistically. Departing from conventional regularization-based or network-expansion-based methods, LPS fully exploits the capacity of the neural network by splitting it into disjoint partitions specialized for each task via pruning; in turn, this mitigates catastrophic forgetting. Simultaneously, to exploit earlier knowledge obtained from previous tasks, LPS  shared parameters between different partitions of the network, in an adaptive, tunable fashion.

\subsection{Architecture Overview}
\begin{figure}[!t]
    \centering
    \includegraphics[width=0.6\columnwidth]{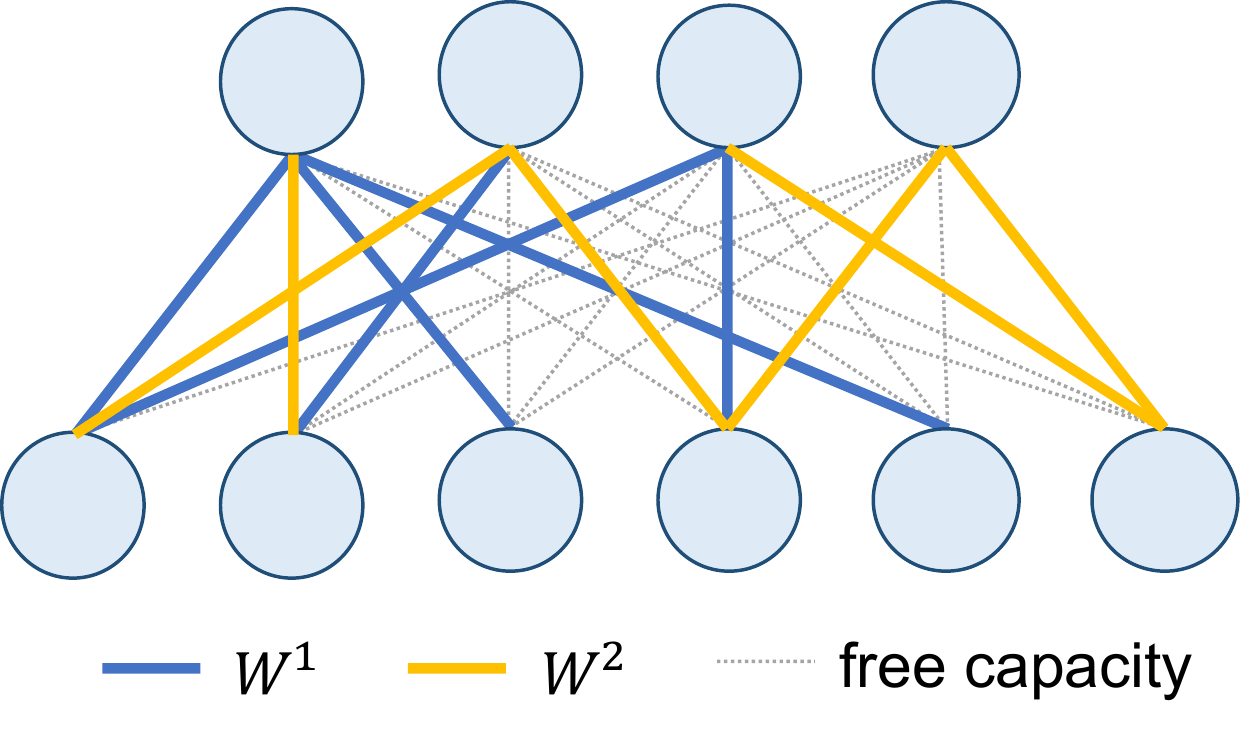}
    \caption{Split of network weights at task 2. Task designated weights $W^1$, $W^2$ have disjoint support, and a lot of excess capacity in the network remains free.}
    \label{fig:network_split}
\end{figure}

\begin{figure*}[!t]
    \centering
    \includegraphics[width=2\columnwidth]{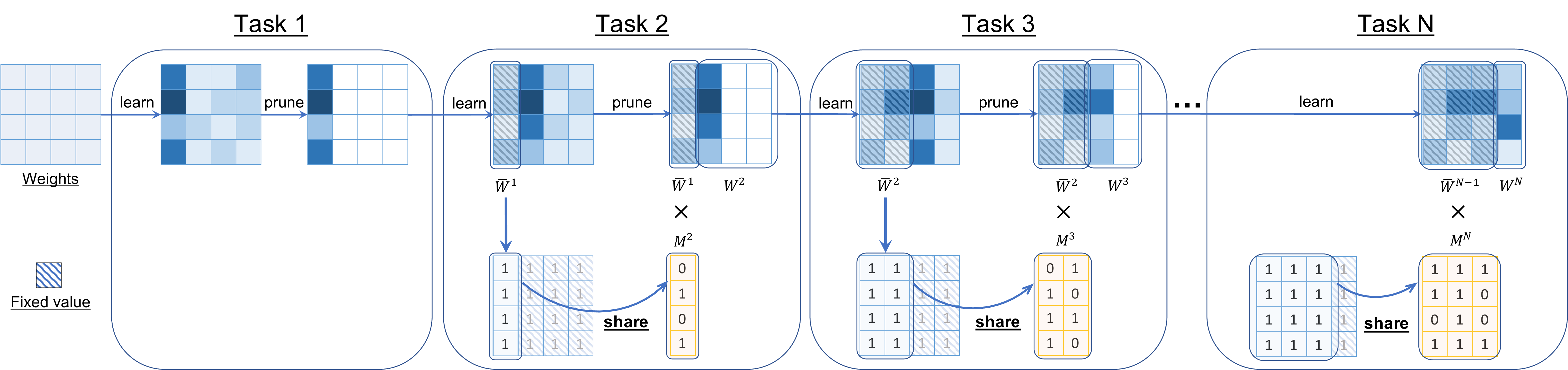}
    \caption{Overview of the proposed LPS method. For each task $t$, given $\bar{W}^{t-1}$ from  previous tasks till $(t−1)$, we \textit{learn} the task, \textit{prune} the neural network to obtain task specific weights $W^t$, and \textit{share} knowledge among tasks via mask $M^t$, simultaneously. Note that for task 1,  we only need to learn $W^1$, as there is no previous knowledge yet; and for the last task N, we do not need to prune unless there is requirement of leaving free capacity for future tasks.}
    \label{fig:overview}
\end{figure*}

We assume that we are given a legacy neural network architecture $f_{W}:{\reals}^d\to {\reals}^{d'}$ (e.g., ResNet \cite{he2016deep}), parameterized by weights $W\in \reals^m$. 
Recall that the support of a vector is the set of its non-zero coordinates. Our solution satisfies the following two  properties: first, at the conclusion of task $t$,  the weights  of the network are partitioned into  task-specific weights $W^1, W^2, \ldots, W^t\in \reals^m$ that have disjoint supports. Formally, for all $1\leq i,j\leq t$  with $i \neq j$:
\begin{align}
    \operatorname{supp} (W^i) &\cap \operatorname{supp} (W^j) = \emptyset.
\end{align}
 Second, these disjoint weights do not exhaust the entire representation capacity of the network: the union of their supports is smaller than $m$.  The remaining weights are treated as excess capacity, to be utilized in future tasks. Formally, let
 \begin{align}
    \bar{W}^t &= \textstyle\sum_{i=1}^{t} W^i\in \reals^m,
\end{align}
 be the sum of the task-specific weights.\footnote{As $W^i$, $i=1,\ldots,t$ have disjoint supports, $\bar{W}^t$ can also be thought of as their superposition.} 
 Then,
 \begin{align}
    \operatorname{supp}(\bar{W}^t)&=\textstyle\bigcup_{i=1}^t\supp(W^i) \leq m.
\end{align}
 Figure \ref{fig:network_split} illustrates the weight split for a single layer at task $t=2$. Weights $\bar{W}^2=W_1+W_2$ are partitioned to two classes $W^1$ and $W^2$ with disjoint support. Moreover,  the excess capacity (the complement of ${\bar{W^2}}$'s support) can be used for future tasks. 

Under this configuration, to make predictions for task $t$, our network  uses $W^t$, i.e. the portion of the network representing task-specific knowledge, as well as \emph{as many of the  weights $\bar{W}^{t-1}$ dedicated to previous tasks as we wish to leverage}. Formally, the network we use for task $t$ has weights
\begin{align}
    W^t + M^t \odot \bar{W}^{t-1},\quad\text{for }t=1,\ldots,n, \label{eq:weights}
\end{align}
where $\odot$ represents element-wise multiplication and $M^t\in \{0,1\}^m$ are a set of learnable \emph{knowledge sharing masks}.

Our solution, and in particular the weight design in Eq.~\eqref{eq:weights},  has several  advantages, each addressing directly the issues of catastrophic forgetting, parsimony,  and knowledge reuse. First,  our approach \emph{does not experience any catastrophic forgetting}. This is precisely because additional tasks are accommodated in excess capacity; classification  for earlier tasks (also through Eq.~\eqref{eq:weights}) remains unaltered. Second, by utilizing only a portion of the overall capacity of the network, we attain parsimony. As we discuss below, this happens at almost no accuracy loss: we learn the small-support parameters $W^i$, $i=1,\ldots,t$ through state-of-the art pruning methods. Finally, the use of masks $M^t\in \{0,1\}^m$ enables arbitrary levels of reuse: setting them to 1 fully reuses weights learned from previous tasks, while setting them to 0 limits the network for task $t$ to only its dedicated weights.  Note that this flexibility comes at the expense of parsimony, as we also need to keep track of masks for each task. As these are binary, however, they are not as memory-intensive as the model weights.  


\subsection{The Learn-Prune-Share (LPS) Algorithm }
Our learn-prune-share algorithm learns task-specific weights $W^t$ as well as knowledge-sharing masks $M^t$ as the datasets $\mathcal{D}_t$ are revealed. It is an iterative process,  summarized in Figure \ref{fig:overview}. At each task, we use the full excess capacity of the network to train a dense network. Using a state-of-the-art pruning method, we reduce this to weights with small support $W^t$; simultaneously, we determine how much of the old weights to reuse via mask $M^t$. This process is repeated until we run out of tasks.

Formally, at each  task $t$, the input to the algorithm consists of (a) earlier weights  from previous tasks $1$ through $(t-1)$, i.e., $\bar{W}^{t-1} = \sum_{i=1}^{t-1} W^i\in \reals^m$, as well as, (b) the  dataset of task $t$ , i.e., $\mathcal{D}^t$. Our goal is  to learn \textit{sparse, small-support} task-specific weights $W^t$, as well as the knowledge-sharing mask $M^t$. Note that for task $1$, we only need to learn $W^1$, as there is no previous knowledge yet. 
As our pruning happens layer-wise, we introduce the following notation. We re-write the weights and masks  as $W = \{ W_l \}_{l=1}^L\in \reals^m$ and $M = \{M_l\}_{l=1}^L\in\{0,1\}^m$  where ${W}_l , M_l$ are the weights and masks, respectively, corresponding to the $l$-th layer, for $l=1,\ldots,L$.
We denote the loss of a  network with weights $W$ under dataset $D$ as $\mathcal{L}(W,W_{L+1}; \mathcal{D})$, where $W_{L+1}\in \reals^{P_{L+1}\times Q_{L+1}}$ is the final (classification) layer. 
In light of Eq.~\eqref{eq:weights}, we formulate the learning process determining $W^t,W_{L+1}^t, M^t$  at task $t$ as an optimization problem:
%
\begin{subequations}
\label{eq:obj}
    \begin{align}
        \underset{W,W_{L+1}, M}{\text{Min.}:}&\quad  \mathcal{L}\left( W + M \odot \bar{W}^{t-1},W_{L+1}; \mathcal{D}^t
        \right), \displaybreak[0]\\
                \label{cons:prune}
        \text{subj.~to:}& \quad W_l \in S_l^t, \quad l = 1, \cdots, L,\displaybreak[0] \\
          & \quad M_l \in S_l^{'t},\quad l = 1, \cdots, L,  \label{cons:mask}\displaybreak[0]\\
        & \quad  \operatorname{supp} (W) \cap \operatorname{supp} (\bar{W}^{t-1}) = \emptyset,\label{cons:non-overlap}\displaybreak[0] \\
       & \quad\operatorname{supp}(M) \subseteq \operatorname{supp}
     (\bar{W}^{t-1}), \label{cons:mbelow} \displaybreak[0]\\
        &  \quad W\in \reals^m, W^t_{L+1}\in \reals^{P_{L+1}\times Q_{L+1}},\\
        & \quad M\in \{0,1\}^m
    \end{align}
\end{subequations}
where $S_l^t$ are \emph{sparsity constraints} on $W_l^t$,  and ${S}_l^{'t}$ are \emph{knowledge-sharing constraints on $M_l^t$}. We describe both in detail below, in Sections \ref{subsec:pruning} and \ref{subsec:sharing}, respectively. 

The constraint in Eq.~\eqref{cons:non-overlap} enforces that weights are indeed disjoint: the weights of $W^t\in\reals^m$ are taken from the current excess capacity pool -- the complement of $\supp(W^{t-1})$. Similarly, the constraint in Eq.~\eqref{cons:mbelow} enforces that the  knowledge-sharing mask $M\in\{0,1\}^m$ are applied to the past weights $W^{t-1}$ only. Note that, together, they imply that $W^{t}$ and $M^t$ have disjoint supports. Finally, the fully connected classifier/output weights $W^{L+1}$ are unconstrained.

\subsection{Task-Specific Weight Constraints}
\label{subsec:pruning}



\begin{figure}[!t]
    \centering
    \scriptsize
    \includegraphics[width=1\columnwidth]{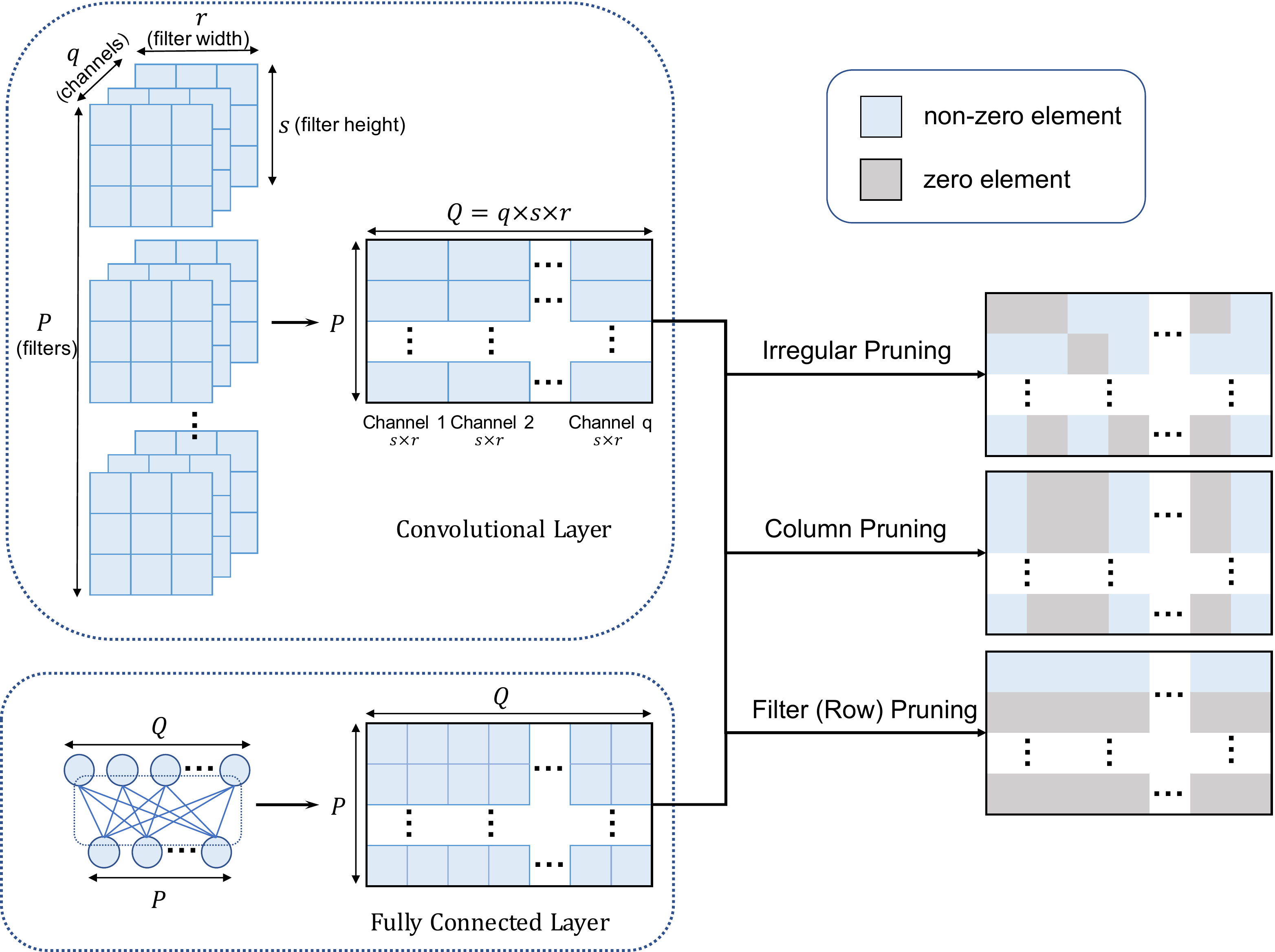} \\
    \caption{Pruning strategy illustration. By converting weights to the format of GEneral Matrix Multiplication operations (GEMMs), we represent both CV and FC layers via the (reshaped) weight matrix $W \in \mathbb{R}^{P \times Q}$. We can then choose from \textbf{\emph{irregular}} or \textbf{\emph{structured}} (i.e. column and filter) pruning.}
    \label{fig:pruning}
\end{figure}
 
To obtain $W^t$, we need to create constraints on $W=\{W_l\}_{l=1}^L\in \reals^m$ in Prob.~\eqref{eq:obj} that enforce sparsity. Recall that we denote the weights of the $l$-th layer of our neural network as $W_l$. In convolutional layers, the weight for $l$-th layer is represented by a four-dimensional tensor, where dimensions $p_l, q_l, r_l, s_l \in \naturals$ correspond to the number of filters, number of channels, filter width, and filter height, respectively. In fully connected layers, weights are $P_l\times Q_l$ matrices, where $P_l$ and $Q_l$  represent the input and output layer size, respectively. We nevertheless assume that all layers are represented in a  GEneral Matrix Multiplication operations (GEMMs) format, which is a standard practice in tensor framework implementations: that is, we assume all tensors are reshaped to two dimensional $P_l\times Q_l$ matrices. This is already the case for fully connected layers; for convolutional layers, the reshaping can take the form $P_l=p_l$ and $Q_l=q_r\cdot r_l \cdot s_l$.   We thus assume  every layer is represented  by a (reshaped) weight matrix $W_l \in \mathbb{R}^{P_l \times Q_l}$, as illustrated in Figure \ref{fig:pruning}. Note that, under this assumption, the total number of weights in the model is $m=\sum_{l=1}^LP_l\cdot Q_l$.

Under this representation, we consider the following sets of constraints $S_l^t$ for  layer $l$:

\noindent\textbf{Irregular Pruning.} For irregular pruning, we have:
 \begin{align}\label{eq:irr}
     S_l^t = \{ {W}_l\in \reals^{P_l\times Q_l}\mid \|{W}_l \|_0 \leq \alpha_l^t \},
 \end{align}
 where $\|\cdot\|_0$ the size of ${W}_l$'s support (i.e., the number of non-zero elements), and $\alpha_l^t\in \mathbb{N}$ is a constant limiting the proportion of non-zero elements. Intuitively, this implies that  $W_l$ has no more than $\alpha_l^t$ non-zero elements.

\noindent\textbf{Structured Pruning.} Given $\phi$ a Boolean predicate, let $\id_{\phi}$ to be 1 if $\phi$ is true, and 0 otherwise. Moreover, given matrix $W_l\in\reals^{P_l\times Q_l}$, let 
$[W_l]_{:,q}\in \reals^{P_l}$ be the 
$q$-th column of $W_l$. 
In \textbf{\emph{column pruning}}, the constraint set $S_l^t$ is defined as:
\begin{align}\label{eq:column}
     S_l^t = \big\{{W}_l\in \reals^{P_l\times Q_l} \mid   \big(\textstyle\sum_{q=1}^{Q_l} \id_{([{W}_l]_{:,q}\neq \mathbf{0})} \big) \leq\alpha_l^t\big\},
\end{align}
where $\alpha_l^t\in \naturals.$  This  enforces that the number of non-zero \emph{columns} in the $l$-th layer's GEMM representation does not exceed $\alpha_l^t$. A similar constraint can be placed on filters/rows of $W_l$ to form structured \textbf{\emph{filter pruning}}, which enforces that the number of non-zero \emph{filters} does not exceed $\alpha_l^t$.

All three types of constraints (irregular,  column, and filter pruning) are illustrated in Fig.~\ref{fig:pruning}. They all lead to disjoint supports if used consistently across tasks: for example, filter pruning ends up partitioning rows of the GEMM representation of every later, column pruning partitions columns, etc., while irregular pruning partitions individual matrix entries.
 
 \subsection{Knowledge-Sharing Mask Constraints}
 \label{subsec:sharing}
 
 %
To control knowledge sharing, we impose a sparsity constraint on $M$ as well, allowing only $\beta_l^t\in \naturals$ of entries in the mask to be non-zero. Formally:  
 \begin{align} \label{eq:share-constraint}
 \begin{split}    S_l^{'t} = \big\{{M}_l\in \{0,1\}^{P_l\times Q_l} \mid   \|{M}_l^t \|_0 = \beta_l^t \big\}.\end{split}
\end{align}
Adjusting the ``sharing parameter'' $\beta_l^t$ allows us to  limit the proportion of old weights shared (i.e., the non-zero elements of $M_l$). 
By forcing ${M}_l$ to be sparse, we force training to select the most beneficial weights for the current task from previously learned weights. Sharing parameter $\beta_l^t$ also conveys the \textit{usefulness} of previous knowledge: e.g. when tasks are similar, previous knowledge would indeed be useful for subsequent tasks, thus $\beta_l^t$ should be big; conversely, for dissimilar tasks we expect fewer sharing opportunities.

\subsection{Solving LPS via ADMM}\label{subsec:ADMM}
The optimization problem defined in Eq.~\eqref{eq:obj} for LPS has non-convex constraints, and solving it via standard stochastic gradient descent is not possible. We use the widely deployed Alternating Direction Method of Multipliers (ADMM)~\cite{boyd2011distributed}, that has been extensively applied in pruning literature \cite{zhang2018systematic,ren2019admm}. For completeness, we describe the ADMM solution to Problem \eqref{eq:obj} in detail in Appendix~\ref{sec:admm-appendix}. In short, ADMM decomposes the original non-convex problem with constraints into subproblems that can be solved separately. It alternates between (a) standard gradient descent with a quadratic proximal penalty (Eq.~\eqref{eq:primal1}), that forces the solution to be close to a value in the (non-convex) constraint space, and (b) an orthogonal projection operation to the constraint space (Eq.~\eqref{eq:admm-process-z}). Hence starting from full weights $W$ and masks $M$ set  to 1, we can progressively prune and constrain both, producing a feasible solution at convergence. Most importantly, the weights and masks can be trained jointly and dynamically. 


From an implementation standpoint, to incorporate our constraints to ADMM, it suffices to produce polynomial-time functions that compute the orthogonal projection into  constraints \eqref{cons:prune} -- \eqref{cons:mask}. 
For \eqref{cons:prune}, polynomial algorithms are well known for  irregular, column, and filter pruning constraints \cite{zhang2018systematic}. For example, for irregular pruning, the orthogonal projection of a matrix $Z\in \reals^{P_l\times Q_l}$ to set $S_l^t$ given by Eq.~\eqref{eq:irr} can be computed by keeping the  $\alpha_l^t$ entries of $Z$ of largest absolute value intact, and setting the rest to zero. For column pruning (Eq.~\eqref{eq:column}), projection of $Z$ to  can $S_l^t$ be computed by similarly keeping the $\alpha_l^t$ columns with largest $\ell_2$ norm intact, and setting all other rows to $\textbf{0}$.

Our mask constraint \eqref{eq:share-constraint} is  more complex, as projection requires not only enforcing sparsity exactly, but also that the values of the matrix become binary. Nevertheless, we can compute the projection of $Z\in\reals^{P_l\times Q_l}$ to $S_l^{'t}$ in polynomial time via the following steps:
\begin{algorithmic}
\STATE Sort elements of matrix $Z$ from smallest to largest;
\STATE Map the largest $\beta_l^t$ entries to 1; set the rest entries to 0.
\end{algorithmic}
We prove the correctness of this algorithm in Appendix~\ref{sec:project-appendix}.




\section{Experiments}
In our experiments, (a) we show that our method outperforms current state-of-the-art methods on both benchmark and real datasets; (b) we assess the importance of the knowledge-sharing mask under different task settings; and (c) we explore how different pruning strategies affect the prediction accuracy.
\subsection{Experimental Setting.}
\noindent\textbf{Datasets.} To evaluate the performance of our approach empirically, we   experiment with two standard lifelong learning benchmark datasets, permuted MNIST \cite{lecun1998mnist, goodfellow2013empirical} and split CIFAR-10/100 \cite{krizhevsky2009learning}, and a real world radiofrequency fingerprinting dataset (split RF) \cite{Jian-iot2020}, summarized in Table \ref{tab:params}.   The original MNIST dataset \cite{lecun1998mnist, goodfellow2013empirical} contains $28 \times 28$ black and white images of handwritten digits of 10 classes. 
Following~\cite{zenke2017continual}, we construct 10 tasks by applying the same random permutation across all MNIST images, using a different permutation for each task. CIFAR-10  \cite{krizhevsky2009learning} comprises 10 classes of 32x32 colour images. 
CIFAR-100 is just like CIFAR-10 in image format and total number of images, but has 100 classes. Following~\cite{zenke2017continual}, we set the first task as the whole CIFAR-10 dataset. We then create 5 additional tasks, each containing 10 consecutive classes from the CIFAR-100 dataset. 
Finally,  the split RF dataset~\cite{Jian-iot2020, gritsenko2019finding} contains radio transmissions from 50 WiFi devices recorded in the wild. 
We randomly partition these 50 classes into 5 tasks. 


\noindent\textbf{Lifelong Learning Methods.}\label{sec:competing}
We compare LPS to the following methods:

\emph{Elastic Weight Consolidation (EWC)  \cite{kirkpatrick2017overcoming}:} EWC applies Laplace Approximation to estimate the importance scores of parameters for previous tasks and uses a quadratic regularizer weighted by the importance scores. 

\emph{Intelligent Synapses (IS) \cite{zenke2017continual}:} IS uses an importance score based regularizer similar to EWC. However, a path integral based method is proposed to evaluate the importance score.

\emph{Learning without Forgetting (LwF) \cite{li2017learning}:} LwF maintains responses for previous tasks via a knowledge distillation loss.

\emph{Deep Generative Replay (DGR) \cite{shin2017continual}:} DGR uses generative adversarial networks (GAN) \cite{goodfellow2014generative} to mimic the data distribution for each task. A generator is updated at every task to incorporate its data distribution. A corresponding classifier is trained using the mixture of generated and new data.

\emph{Gradient Episodic Memory (GEM) \cite{lopez2017gradient}:} GEM proposes an episodic memory saving a portion of previous data and use the loss on this data a constraint when training a new task.

\emph{PackNet \cite{mallya2018packnet}:} PackNet iteratively prunes the model to accommodate new tasks by removing parameters of smaller magnitude heuristically. Similar formulation is proposed by \cite{CLNP} under a lifelong learning setting. 

We use the implementation from the original authors for all methods, including the recommended hyperparameter settings or tuning strategies. 
The same network architectures are used among all methods for fair comparison.

\begin{table}
    \centering
    \scriptsize
    \setlength{\extrarowheight}{.2em}
    \setlength{\tabcolsep}{3.2pt}
    \caption{Dataset and Parameter Summary.}
    \label{tab:params}
    \begin{tabular}{||c c ||c|c c|c||}
        \hline
        \multicolumn{2}{||c||}{\multirow{3}{*}{Stat. \& Param.}} & \multicolumn{4}{c||}{Datasets} \\
        \cline{3-6}
        & & Permuted MNIST & \multicolumn{2}{c|}{Split CIFAR} & Split RF \\
            & &  & $n=1$ & $n\neq 1$ & \\
        \hline
        \hline
        \multicolumn{2}{||c||}{\# tasks ($n$)} & 10 & \multicolumn{2}{c|}{6} & 5 \\
        
        \multicolumn{2}{||c||}{\# classes per task} & 10 & \multicolumn{2}{c|}{10} & 10 \\

        \multicolumn{2}{||c||}{\# train samples per task}
        & 60,000 & 50,000 & 5,000 & 1,410 \\

        \multicolumn{2}{||c||}{\# test samples per task}
        & 10,000 & 10,000 & 1,000 & 550 \\
        \hline
        \hline
        \multicolumn{2}{||c||}{{$\alpha^t_l$} (\% total layer params)}
        & 10\% & 50\% & 10\% & 20\%\\

        \multicolumn{2}{||c||}{{$\beta^t_l$} (\% total $\bar{W}^{t-1}_l$ params)} 
        & 90\% & \multicolumn{2}{c|}{92\%} & 90\%\\

        \multicolumn{2}{||c||}{Pruning strategy}
        & Irregular & \multicolumn{2}{c|}{Irregular} & Column \\
        \hline
        \multicolumn{2}{||c||}{LPS Epochs} 
        & \multirow{2}{*}{30/90/30} & \multicolumn{2}{c|}{\multirow{2}{*}{200/600/200}} & \multirow{2}{*}{20/60/20} \\
        \multicolumn{2}{||c||}{(warm-up/ADMM/final)}   &  & \multicolumn{2}{c|}{} & \\
        \hline
              \hline
        \multicolumn{2}{||c||}{Architecture} & Two FC layers & \multicolumn{2}{c|}{CIFAR-10} & ResNet50-1D \\
        \multicolumn{2}{||c||}{\# params ($m$)} & 5,568,000 & \multicolumn{2}{c|}{884,576} & 15,901,568 \\
        \multicolumn{2}{||c||}{\# layers ($L$)} & 2 & \multicolumn{2}{c|}{5} & 49 \\
        
        \hline
    \end{tabular}
\end{table}

\noindent\textbf{Architectures.} 
We implement different architectures for permuted MNIST, split CIFAR-10/100, and split RF, respectively. The architecture for permuted MNIST dataset~\cite{zenke2017continual} contains two hidden layers, each with 2000 neurons and ReLU activations. For split CIFAR-10/100 dataset, we use the default
CIFAR-10 architecture from \href{https://raw.githubusercontent.com/fchollet/keras/keras-2/examples/cifar10_cnn.py}{Keras}~\cite{zenke2017continual}. For split RF dataset, we use ResNet50-1D~\cite{resnet}, which is the 1D-convolutional version of ResNet50, targeting inputs as 2D fixed-length sequences. 
For all three architectures, we learn the biases and batch normalization parameters for the first task and keep these terms fixed for subsequent tasks. 

\noindent\textbf{LPS Implementation.}
For each task, we run LPS in three phases. In the warm-up phase, we first train a $W$ over the full free parameters with $M=1$. In the ADMM phase, we then prune the network Eq.~\eqref{eq:admm-process}. In the final stage, we do a final projection to the constraint sets of both masks and weights, and retrain the weights, changing only non-zero values. 
We set all $\rho_i=10^{-3}$ and increase by a factor of 10 at equal intervals during ADMM iterations. 
We use the following hyperparameters, which we  determine using a validation set.  
Unless otherwise noted,   sparsity parameters $\alpha_l^t$ and $\beta_l^t$ are as shown in Table~\ref{tab:params}. We explored the impact of both in \Cref{sec:experiments}. 
For all experiments, we use a batch size of 128 and Adam~\cite{adam} as an optimizer with default values and initialize the learning rate to 0.001. Our proposed LPS approach is implemented in Python using PyTorch~\cite{pytorch} and NVIDIA CUDA support. 
All experiments are carried out on an  Tesla V100 GPU with 32 GB memory and 5120 cores.

\noindent\textbf{Evaluation Metrics.}
 We evaluate the final obtained model (associated with masks and multi-head output layers) on all tasks testsets via (Top-1) accuracy.

\subsection{Results on Benchmark Datasets}\label{sec:experiments}

\paragraph{Effectiveness of the proposed LPS approach.}
Table~\ref{tab:overall} shows the overall performance, in terms of the final average accuracy across all tasks, of all lifelong learning methods. For reference purposes, we also include the accuracy attained when training a full-capacity (non-parsimonious) single model separately for each task (SM).  LPS outperforms all  competitors across all datasets. Most methods perform well on permuted MNIST; the margin is wider on the remaining two datasets, that are more challenging. To further scrutinize the performance of LPS across tasks, we show in   
Table~\ref{tab:baseline-cifar}-\ref{tab:baseline-rf} the per task accuracy. Interestingly LPS outperforms all competitors across all tasks on both datasets; we also observed this on the 10 tasks of the permuted MNIST, which we omit for brevity. Overall, \emph{our LPS approach achieves both the best average and the best task-specific accuracy for all three datasets}. 


We further observe that regularization-based methods like EWC and IS perform relatively well on benchmarks, while they fail on split RF. One possible explanation may be that when tasks are more diverse and model is large, regularizers do not suffice to keep the learned information. Evidence of forgetting is present in LwF, for split CIFAR, and almost all methods (except LPS and PackNet) on split RF. This is expected, as both LPS and PackNet are immune to  forgetting. 

We also observe that LSP even outperforms the full-capacity SM trained from scratch on each task for split CIFAR-10/100 and split RF, and is very close to it over permuted MNIST. This happens despite the fact that it uses only a small fraction of the $m$ parameters used by SM, indicating that it avoids overfitting. Also, we see a clear benefit of reuse of parameters across tasks in split CIFAR (Table~\ref{tab:baseline-cifar}): by partially utilizing past weights, prediction on later tasks improves under LPS compared to SM. 

\begin{table}[!t]
    \centering
    \setlength{\extrarowheight}{.2em}
    \setlength{\tabcolsep}{2.65pt}
    \scriptsize
    \caption{Overall performance on three benchmark datasets. For all the methods, we report the final average accuracy (\%) across all tasks. We include SM (column 2) for reference purpose, which trains a full-capacity single model separately for each task. LPS parameters are set as in Table~\ref{tab:params}.} 
    \label{tab:overall}
    \begin{tabular}{||c ||c||cccccc|c||}
        \hline
        \multirow{2}{*}{Datasets} & \multicolumn{8}{c||}{Methods} \\
        \cline{2-9}
        & SM & EWC & IS & LwF & DGR & GEM & PackNet & LPS \\
        \hline
        \hline
        Permuted MNIST     & 98.80 & 96.81 & 97.52 & 68.22 & 90.73 & 93.03 & 98.14 & \textbf{98.58} \\
        Split CIFAR-10/100 & 75.14 & 71.13 & 74.97 & 54.68 & 63.61 & 66.05 & 77.79 & \textbf{80.13} \\
        Split RF           & 81.15 & 37.01 & 42.63 & 27.75 & 48.27 & 68.38 & 79.37 & \textbf{81.22} \\
        \hline
    \end{tabular}
\end{table}

\begin{table}[!t]
    \centering
    \setlength{\extrarowheight}{.2em}
    \scriptsize
    \caption{Split CIFAR-10/100: For all the methods, we report the task-specific, and the final average accuracy (\%) across all tasks. LPS parameters are set as in Table~\ref{tab:overall}.}
    \label{tab:baseline-cifar}
    \begin{tabular}{||c ||c|c|c|c|c|c||c||}
        \hline
        \multirow{2}{*}{Methods} & \multicolumn{7}{c||}{Tasks} \\
        \cline{2-8}
        & task 1 & task 2 & task 3 & task 4 & task5 & task 6 & Avg. \\
        \hline
        \hline
        SM & 82.32 & 75.40 & 70.20 & 75.90 & 71.70 & 75.30 & 75.14\\
        \hline
        \hline
        EWC & 71.23 & 72.50 & 69.25 & 71.34 & 67.52 & 74.93 & 71.13\\
        IS & 74.59 & 74.28 & 74.19 & 75.54 & 75.58 & 75.62 & 74.97\\
        LwF & 40.32&	56.77&	48.60&	53.94&	60.04&	68.43& 54.68\\
        DGR & 64.36&	62.01&	63.02&	67.34&	65.28&	59.64& 63.61\\
        GEM & 68.52&	65.34&	63.88&	70.12&	65.23&	63.23& 66.05\\
        PackNet & 82.33 & 79.30 & 73.90 & 78.80 & 74.30 & 78.10 & 77.79 \\
        
        \hline
        LPS & \textbf{82.97} & \textbf{80.00} & \textbf{76.50} & \textbf{79.90} & \textbf{78.40} & \textbf{83.00} & \textbf{80.13} \\
        \hline
    \end{tabular}
\end{table}

\begin{table}[!t]
    \centering
    \scriptsize
    \setlength{\extrarowheight}{.2em}
    \setlength{\tabcolsep}{8pt}
    \caption{Split RF: For all the methods, we report the task-specific, and the final average accuracy (\%) across all tasks. LPS parameters are set as in Table~\ref{tab:overall}.}
    \label{tab:baseline-rf}
    \begin{tabular}{||c ||c|c|c|c|c||c||}
        \hline
        \multirow{2}{*}{Methods} & \multicolumn{6}{c||}{Tasks} \\
        \cline{2-7}
        & task 1 & task 2 & task 3 & task 4 & task 5 & Avg. \\
        \hline
        \hline
        SM & 76.33 & 73.50 & 85.30 & 85.60 & 85.00 & 81.15 \\
        \hline
        \hline
        EWC & 25.73 & 35.32 & 30.85 & 45.81 & 47.24 & 37.01\\
        IS & 27.08 & 40.72 & 37.25 & 50.66 & 57.34& 42.63\\
        LwF & 14.62	& 20.37	& 23.45	& 33.58	& 46.72	& 27.75 \\
        DGR & 43.50	& 49.37	& 43.87	& 50.25	& 54.38	& 48.27 \\
        GEM & 67.24	& 63.45	& 68.53	& 70.26	& 72.44	& 68.38 \\
        PackNet & 78.15 & 74.14 & 82.56 & 80.54 & 81.45 & 79.37 \\
        
        \hline
        LPS & \textbf{78.33} & \textbf{77.55} & \textbf{84.19} & \textbf{82.63} & \textbf{83.39} & \textbf{81.22} \\
        \hline
    \end{tabular}
\end{table}

\begin{figure*}[!t]
	\centering
	\includegraphics[width=1.95\columnwidth]{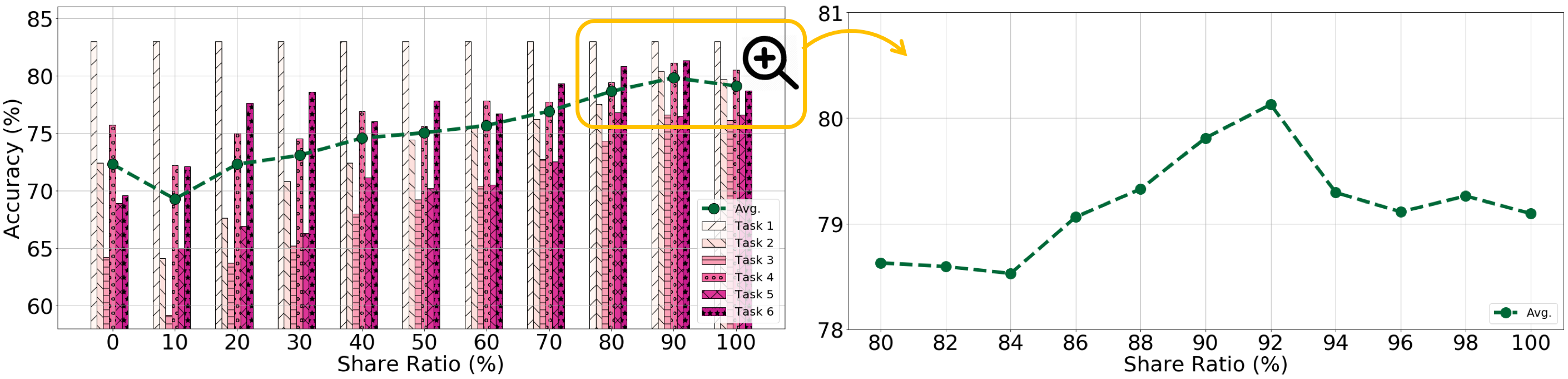}\\
    \caption{Split CIFAR-10/100: An exploration of how average and per task accuracy changes as we modify the share ratio $\beta^t_l$. The x-axis is the share ratio, indicating the ratio of the parameter over the total number of past weights per layer. For each share ratio, we represent the task specific and average accuracy as the colored bar and the green dot, respectively. The optimal  is at 92\% share, depicted right.}
	\label{fig:cifar-share}
\end{figure*} 

\begin{figure*}[!t]
	\centering
	\includegraphics[width=1.95\columnwidth]{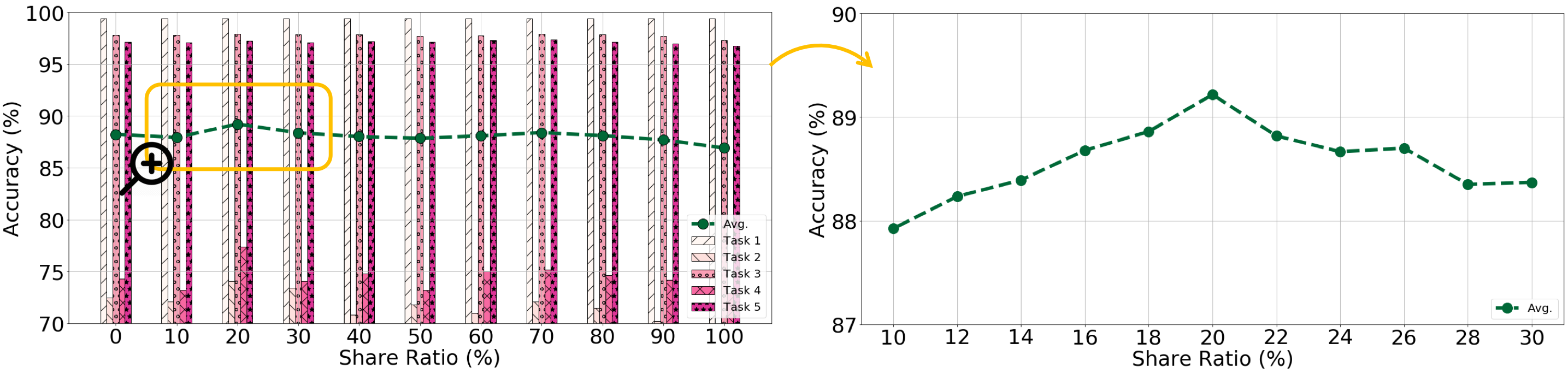}\\
	\caption{Mixed Dataset: An exploration of how average and per task accuracy changes as we modify the share ratio $\beta^t_l$ on non-similar tasks. The optimal value is at 20\% share, depicted in detail in the right figure. Less knowledge reuse performs even better, demonstrating LPS does adaptively select useful knowledge for the current task, and indicating the share strategy choice should be guided by the inter-task similarity.
	}
	\label{fig:mixture-share}
\end{figure*} 

\begin{table*}
    \centering
    \setlength{\extrarowheight}{.2em}
    \caption{LPS with no (0\%), full (100\%) and selective share on three benchmark datasets. For selective share, we follow the same parameter  searching  strategy  as  in  split CIFAR-10/100  to  get the best performing model. To make a fair comparison, we start experiments from the learned model on task 1 (no previous knowledge yet), then sequentially train this model on remaining tasks with different share ratio $\beta^t_l$.}
    \label{tab:share-ratio}
    \begin{tabular}{||c c||c|c|c|c|c|c|c|c|c|c||c||}
        \hline
        \multirow{2}{*}{Datasets} & \multirow{2}{*}{$\beta^t_l$} & \multicolumn{11}{c||}{Tasks} \\
        \cline{3-13}
        & & task 1 & task 2 & task 3 & task 4 & task5 & task 6 & task 7 & task 8 & task 9 & task 10 & Avg. \\
        \hline
        \hline
        \multirow{3}{*}{Permuted  MNIST}
        & 0\% & \textbf{98.92} & \textbf{98.77} & 98.47 & 98.51 & \textbf{98.58} & 98.49 & 98.29 & 97.91 & 97.78 & 85.82 & 97.15 \\
        & 100\% & \textbf{98.92} & 98.56 & 98.51 & 98.39 & 98.35 & 98.24 & 98.26 & 98.19 & 98.25 & 98.14 & 98.38 \\
        & 90\% & \textbf{98.92} & 98.68 & \textbf{98.71} & \textbf{98.64} & 98.55 & \textbf{98.61} & \textbf{98.49} & \textbf{98.51} & \textbf{98.42} & \textbf{98.23} & \textbf{98.58} \\
        
        \hline
        \hline
        \multirow{3}{*}{Split CIFAR-10/100}
        & 0\% & \textbf{82.97} & 72.40 & 64.20 & 75.70 & 68.90 & 69.60 &\cellcolor{Color1}-&\cellcolor{Color1}-&\cellcolor{Color1}-&\cellcolor{Color1}-& 72.30 \\
        & 100\% & \textbf{82.97} & 79.70 & 76.10 & \textbf{80.50} & 76.60 & 78.70  &\cellcolor{Color1}-&\cellcolor{Color1}-&\cellcolor{Color1}-&\cellcolor{Color1}-& 79.10 \\
        & 92\% & \textbf{82.97} & \textbf{80.00} & \textbf{76.50} & 79.90 & \textbf{78.40} & \textbf{83.00} &\cellcolor{Color1}-&\cellcolor{Color1}-&\cellcolor{Color1}-&\cellcolor{Color1}-& \textbf{80.13} \\
        
        \hline
        \hline
        
        \multirow{3}{*}{Split RF}
        & 0\% & \textbf{78.33} & 77.33 & 83.29 & 81.90 & 82.20 &\cellcolor{Color1}-&\cellcolor{Color1}-&\cellcolor{Color1}-&\cellcolor{Color1}-&\cellcolor{Color1}-& 80.61 \\
        & 100\% & \textbf{78.33} & \textbf{77.59} & \textbf{84.93} & 81.90 & 83.12 &\cellcolor{Color1}-&\cellcolor{Color1}-&\cellcolor{Color1}-&\cellcolor{Color1}-&\cellcolor{Color1}-& 81.17 \\
        & 90\% & \textbf{78.33} & 77.55 & 84.19 & \textbf{82.63} & \textbf{83.39} &\cellcolor{Color1}-&\cellcolor{Color1}-&\cellcolor{Color1}-&\cellcolor{Color1}-&\cellcolor{Color1}-& \textbf{81.22} \\
        \hline
    \end{tabular}
\end{table*}

\paragraph{Share Parameter Effects.} We further explore the impact of knowledge-sharing in Figure \ref{fig:cifar-share}. The figure shows how average and per task accuracy changes as we modify $\beta^t_l$: the $x$-axis is the share ratio, indicating the ratio of the parameter over the total number of past weights per layer on the CIFAR dataset. The optimal value is at 92\%. Moreover, we clearly see that a large reduction in sharing has a bigger impact on later tasks-which otherwise would benefit from knowledge reuse. 

We also show the results of models with no (0\%) and full (100\%) share on all datasets as well as our best performing model with selective sharing in Table \ref{tab:share-ratio}. We follow the same parameter searching strategy as in split CIFAR-10/100 to get the best performing model on validation set. Interestingly, for all three datasets, we observe the best performance achieved by setting share ratio around 90\%. This also indicates that many (not all) past weights  are valuable or meaningful for new tasks. 

To explore this notion of knowledge re-use further, we conducted an experiment in which tasks vary drastically. To do so, we construct a 5-task ``mixed'' dataset,  where tasks 1,3,5 are from the MNIST dataset, with different permutation patterns and tasks 2, 4 both contain 10 different classes from CIFAR-100. Images from permuted MNIST are augmented to RGB images by repeating 3 channels using the original image and resized to $32 \times 32$ to be compatible with CIFAR images.
Similar to Figure \ref{fig:cifar-share}, Figure \ref{fig:mixture-share} shows the effect of the sharing ratio on the mixture dataset. Not surprisingly, the behavior is quite different from Fig.~\ref{fig:cifar-share}. The highest accuracy (89.22\%) is achieved by 20\% share, which demonstrates that LPS does adaptively select useful knowledge for the current task. Note that, faced with these dissimilar tasks, full share (88.15\%) performs even worse than no share (88.23\%), indicating the share strategy choice should be flexible and guided by the inter-task similarity.


\paragraph{Comparing different pruning strategies.}
We compared three different pruning strategies (i.e., column, filter, and irregular pruning) on split CIFAR-10/100 and split RF datasets, summarized in Table \ref{tab:structured-cifar} and Table \ref{tab:structed-rf}, respectively. Both irregular and column pruning obtain satisfactory performance, achieving 80.13\% and 79.56\% on split CIFAR-10/100, 80.55\% and 81.22\% on split RF, respectively. However, filter pruning reflects an unstable performance, obtaining 68.11\% and 80.12\% on split CIFAR-10/100 and split RF datasets, respectively.

\begin{figure}[!t]
	\centering
	\includegraphics[width=.95\columnwidth]{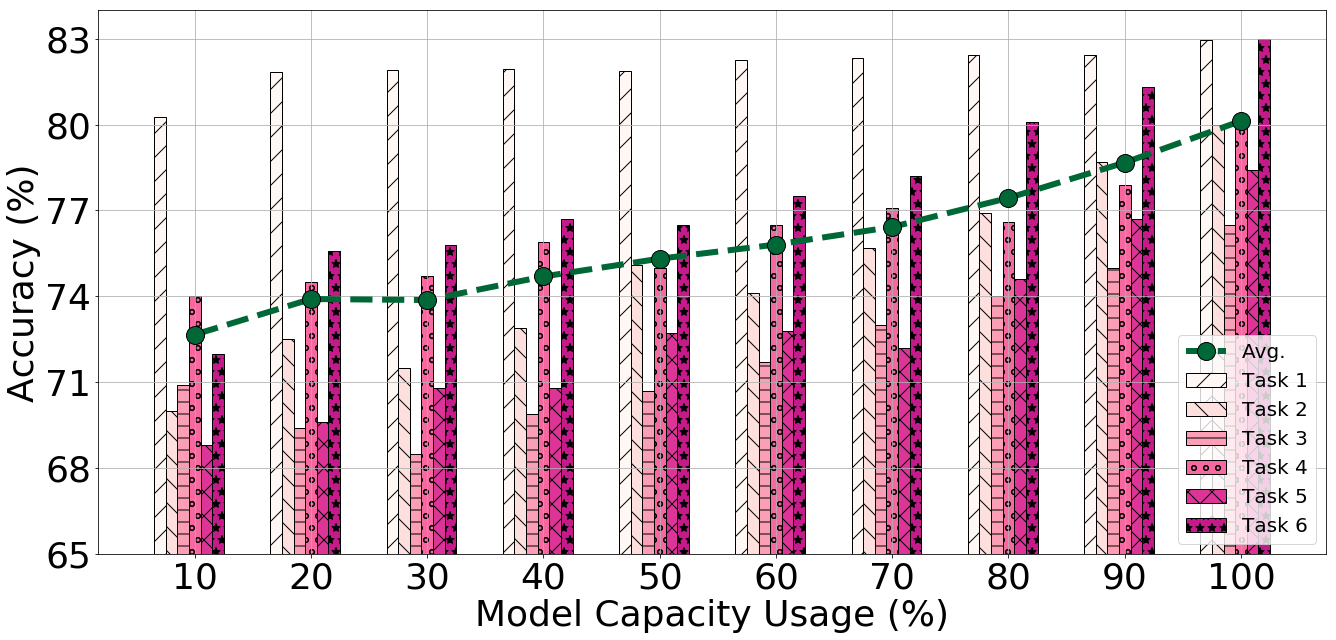}\\
    \caption{Split CIFAR-10/100: To demonstrate our LPS method has the potential capacity to scale to more future tasks, we use only a certain fraction (e.g., $x\%$) of the full model by the 6-th task, leaving $1-x\%$ parameters free for future growth. The x-axis is the fraction of the model capacity usage. As it can be observed, LPS achieves 75.32\% average accuracy with only 50\% model capacity, which is even better than the best non-pruning method IS (74.97\%) with full model capacity.}
	\label{fig:cifar-capasity}
\end{figure}

\paragraph{Impact of Model Capacity.}
Figure \ref{fig:cifar-capasity} measures how model capacity usage affects the accuracy on the split CIFAR-10/100 dataset. For this experiment, instead of using the whole model capacity for the 6 tasks, we use only a fraction  (e.g., $x\%$) of the full model by the $n$-th task, leaving $1-x\%$ parameters free for future growth; all other parameters are set as in Table~\ref{tab:params}. Figure \ref{fig:cifar-capasity} shows the impact on average and per task accuracy as we vary fraction $x$. We clearly observe that a model performs better when more capacity is available. Nevertheless, accuracy performance is also robust to this shrinkage --  it achieves 75.32\% accuracy with only 50\% model capacity, which is even better than the best non-pruning method IS (74.97\%) with full model capacity. Surprisingly, at only 10\% of the total capacity of the network, accuracy does not collapse, but still remains above 72.5\%. This indicates that our method has the potential capacity to scale to even more future tasks. 





\begin{table}[!t]
    \centering
    \scriptsize
    \setlength{\extrarowheight}{.2em}
    \setlength{\tabcolsep}{4.5pt}
    \caption{Three pruning strategies on split CIFAR-10/100.}
    \label{tab:structured-cifar}
    \begin{tabular}{||c| c ||c|c|c|c|c|c||c||}
        \hline
        \multirow{2}{*}{\thead{Prun.\\ Appr.}} & \multirow{2}{*}{$\beta^t_l$} & \multicolumn{7}{c||}{Tasks} \\
        \cline{3-9}
        & & task 1 & task 2 & task 3 & task 4 & task5 & task 6 & Avg. \\
        \hline
        \hline
        \multicolumn{2}{||c||}{SM} & 82.32 & 75.40 & 70.20 & 75.90 & 71.70 & 75.30 & 75.14\\
        \hline
        \hline
        \multirow{4}{*}{Irregular} 
        & 0\% & \textbf{82.97} & 72.40 & 64.20 & 75.70 & 68.90 & 69.60 & 72.30\\
        & 100\% & \textbf{82.97} & 79.70 & 76.10 & \textbf{80.50} & 76.60 & 78.70 & 79.10 \\
        & 92\%    & \textbf{82.97} & \textbf{80.00} & \textbf{76.50} & 79.90 & \textbf{78.40} & \textbf{83.00} & \textbf{80.13} \\
        \hline
        \hline
        \multirow{4}{*}{Column} 
        & 0\% & \textbf{82.04} & 68.80 & 56.50 & 71.00 & 63.90 & 63.00 & 67.54\\
        & 100\%    & \textbf{82.04} & 80.80 & 76.20 & 80.30 & 76.40 & 77.90 & 78.94 \\
        & 92\%  & \textbf{82.04} & \textbf{80.90} & \textbf{76.30} & \textbf{80.60} & \textbf{77.10} & \textbf{80.40} & \textbf{79.56} \\
        \hline
        \hline
        \multirow{4}{*}{Filter} 
        & 0\% & \textbf{79.95} & 56.50 & 50.40 & 62.20 & 54.60 & 55.80 & 59.91\\
        & 100\%    & \textbf{79.95} & 60.20 & 60.00 & 60.40 & 58.90 & 61.10 & 63.43 \\
        & 92\%  & \textbf{79.95} & \textbf{62.10} & \textbf{61.70} & \textbf{67.70} & \textbf{66.50} & \textbf{70.70} & \textbf{68.11} \\
        \hline
    \end{tabular}
\end{table}

\begin{table}[!t]
    \centering
    \scriptsize
    \setlength{\extrarowheight}{.2em}
    \setlength{\tabcolsep}{6.1pt}
    \caption{Three pruning strategies on split RF.}
    \label{tab:structed-rf}
    \begin{tabular}{||c| c ||c|c|c|c|c||c||}
        \hline
        \multirow{2}{*}{\thead{Prun.\\ Appr.}} & \multirow{2}{*}{$\beta^t_l$} & \multicolumn{6}{c||}{Tasks} \\
        \cline{3-8}
        & & task 1 & task 2 & task 3 & task 4 & task 5 & Avg. \\
        \hline
        \hline
        \multicolumn{2}{||c||}{SM} & 76.33 &  73.50 &  85.30 &  85.60 &  85.00 &  81.15 \\
        \hline
        \hline
        \multirow{4}{*}{Irregular} 
        & 0\% & \textbf{78.33} &  \textbf{75.14} &  83.74 &  82.19 &  73.03 &  78.49 \\
        & 100\%    & \textbf{78.33} &  \textbf{75.14} &  84.01 &  79.71 &  82.20 &  79.88 \\
        & 90\%    & \textbf{78.33} &  74.21 &  \textbf{84.56} &  \textbf{83.00} &  \textbf{82.65}&  \textbf{80.55} \\
        \hline
        \hline
        \multirow{4}{*}{Column} 
        & 0\% & \textbf{78.33} &  77.33 &  83.29 &  81.90 &  82.20 &  80.61\\
        & 100\% & \textbf{78.33} &  \textbf{77.59} &  \textbf{84.93} &  81.90 &  83.12 &  81.17 \\
        & 90\% & \textbf{78.33} &  77.55 &  84.19 &  \textbf{82.63} &  \textbf{83.39} &  \textbf{81.22} \\
        \hline
        \hline
        \multirow{4}{*}{Filter} 
        & 0\% & \textbf{77.59} &  70.32 &  82.64 &  80.36 &  82.39 &  78.66 \\
        & 100\% & \textbf{77.59} &  73.65 &  82.90 &  80.44 &  82.85 &  79.49 \\
        & 90\% & \textbf{77.59} &  \textbf{74.54} &  \textbf{83.64} &  \textbf{81.72} &  \textbf{83.12} &  \textbf{80.12} \\
        \hline
    \end{tabular}
\end{table}


\section{Conclusions and Future Work}
In this paper, we propose the learn-prune-share (LSP) algorithm for lifelong learning. Our method maintains a parsimonious neural network model and achieves exact no forgetting by splitting the network into task-specific partitions via ADMM-based pruning method. Moreover, a novel selective knowledge sharing scheme is integrated seamlessly into the ADMM optimization framework to address knowledge reuse. Experiments on permuted MNIST, split CIFAR10/100 and split RF demonstrates our approach achieves significant improvement over the state-of-the-art methods. Future directions include applying more advanced pruning strategies on the lifelong learning problem and exploring how to measure the capacity of a model quantitatively.

\section{Acknowledgements}
The authors gratefully acknowledge support by the National Science Foundation (grant CCF-1937500).

\bibliographystyle{IEEEtran}
\bibliography{IEEEabrv,ref}{}

\begin{thebibliography}{10}
\providecommand{\url}[1]{#1}
\csname url@samestyle\endcsname
\providecommand{\newblock}{\relax}
\providecommand{\bibinfo}[2]{#2}
\providecommand{\BIBentrySTDinterwordspacing}{\spaceskip=0pt\relax}
\providecommand{\BIBentryALTinterwordstretchfactor}{4}
\providecommand{\BIBentryALTinterwordspacing}{\spaceskip=\fontdimen2\font plus
\BIBentryALTinterwordstretchfactor\fontdimen3\font minus
  \fontdimen4\font\relax}
\providecommand{\BIBforeignlanguage}[2]{{%
\expandafter\ifx\csname l@#1\endcsname\relax
\typeout{** WARNING: IEEEtran.bst: No hyphenation pattern has been}%
\typeout{** loaded for the language `#1'. Using the pattern for}%
\typeout{** the default language instead.}%
\else
\language=\csname l@#1\endcsname
\fi
#2}}
\providecommand{\BIBdecl}{\relax}
\BIBdecl
\renewcommand{\BIBentryALTinterwordstretchfactor}{4}

\bibitem{thrun1995lifelong}
S.~Thrun and T.~M. Mitchell, ``Lifelong robot learning,'' \emph{Robotics and
  autonomous systems}, vol.~15, no. 1-2, pp. 25--46, 1995.

\bibitem{mccloskey1989catastrophic}
M.~McCloskey and N.~J. Cohen, ``Catastrophic interference in connectionist
  networks: The sequential learning problem,'' in \emph{Psychology of learning
  and motivation}.\hskip 1em plus 0.5em minus 0.4em\relax Elsevier, 1989,
  vol.~24, pp. 109--165.

\bibitem{ratcliff1990connectionist}
R.~Ratcliff, ``Connectionist models of recognition memory: constraints imposed
  by learning and forgetting functions.'' \emph{Psychological review}, vol.~97,
  no.~2, p. 285, 1990.

\bibitem{parisi2019continual}
G.~I. Parisi, R.~Kemker, J.~L. Part, C.~Kanan, and S.~Wermter, ``Continual
  lifelong learning with neural networks: A review,'' \emph{Neural Networks},
  2019.

\bibitem{kirkpatrick2017overcoming}
J.~Kirkpatrick \emph{et~al.}, ``Overcoming catastrophic forgetting in neural
  networks,'' \emph{Proceedings of the national academy of sciences}, vol. 114,
  no.~13, pp. 3521--3526, 2017.

\bibitem{zenke2017continual}
F.~Zenke, B.~Poole, and S.~Ganguli, ``Continual learning through synaptic
  intelligence,'' in \emph{ICML}, 2017, pp. 3987--3995.

\bibitem{li2017learning}
Z.~Li and D.~Hoiem, ``Learning without forgetting,'' \emph{IEEE transactions on
  pattern analysis and machine intelligence}, vol.~40, no.~12, pp. 2935--2947,
  2017.

\bibitem{nguyen2017variational}
C.~V. Nguyen, Y.~Li, T.~D. Bui, and R.~E. Turner, ``Variational continual
  learning,'' in \emph{ICLR}, 2018.

\bibitem{aljundi2018memory}
R.~Aljundi, F.~Babiloni, M.~Elhoseiny, M.~Rohrbach, and T.~Tuytelaars, ``Memory
  aware synapses: Learning what (not) to forget,'' in \emph{ECCV}, 2018, pp.
  139--154.

\bibitem{shin2017continual}
H.~Shin, J.~K. Lee, J.~Kim, and J.~Kim, ``Continual learning with deep
  generative replay,'' in \emph{NeurIPS}, 2017, pp. 2990--2999.

\bibitem{lopez2017gradient}
D.~Lopez-Paz and M.~Ranzato, ``Gradient episodic memory for continual
  learning,'' in \emph{NeurIPS}, 2017, pp. 6467--6476.

\bibitem{van2018generative}
G.~M. van~de Ven and A.~S. Tolias, ``Generative replay with feedback
  connections as a general strategy for continual learning,'' in \emph{COSYNE
  Workshop}, 2019.

\bibitem{ostapenko2018learning}
O.~{Ostapenko}, M.~{Puscas}, T.~{Klein}, P.~{Jähnichen}, and M.~{Nabi},
  ``Learning to remember: A synaptic plasticity driven framework for continual
  learning,'' in \emph{CVPR}, 2019.

\bibitem{rusu2016progressive}
A.~A. Rusu, N.~C. Rabinowitz, G.~Desjardins, H.~Soyer, J.~Kirkpatrick,
  K.~Kavukcuoglu, R.~Pascanu, and R.~Hadsell, ``Progressive neural networks,''
  \emph{arXiv preprint arXiv:1606.04671}, 2016.

\bibitem{yoon2017lifelong}
J.~Yoon, E.~Yang, J.~Lee, and S.~J. Hwang, ``Lifelong learning with dynamically
  expandable networks,'' in \emph{ICLR}, 2018.

\bibitem{draelos2017neurogenesis}
T.~J. Draelos, N.~E. Miner, C.~C. Lamb, J.~A. Cox, C.~M. Vineyard, K.~D.
  Carlson, W.~M. Severa, C.~D. James, and J.~B. Aimone, ``Neurogenesis deep
  learning: Extending deep networks to accommodate new classes,'' in
  \emph{IJCNN}, 2017, pp. 526--533.

\bibitem{CLNP}
S.~Golkar, M.~Kagan, and K.~Cho, ``Continual learning via neural pruning,''
  \emph{arXiv preprint arXiv:1903.04476}, 2019.

\bibitem{mallya2018packnet}
A.~Mallya and S.~Lazebnik, ``Packnet: Adding multiple tasks to a single network
  by iterative pruning,'' in \emph{CVPR}, 2018, pp. 7765--7773.

\bibitem{golkar2019continual}
S.~Golkar, M.~Kagan, and K.~Cho, ``Continual learning via neural pruning,''
  \emph{arXiv preprint arXiv:1903.04476}, 2019.

\bibitem{ye2018progressive-pruning}
S.~Ye \emph{et~al.}, ``Progressive weight pruning of deep neural networks using
  admm,'' \emph{arXiv preprint arXiv:1810.07378}, 2018.

\bibitem{han2015deep}
S.~Han, H.~Mao, and W.~J. Dally, ``Deep compression: Compressing deep neural
  networks with pruning, trained quantization and huffman coding,'' in
  \emph{ICLR}, 2016.

\bibitem{guo2016dynamic}
Y.~Guo, A.~Yao, and Y.~Chen, ``Dynamic network surgery for efficient dnns,'' in
  \emph{NeurIPS}, 2016, pp. 1379--1387.

\bibitem{dong2019network}
X.~Dong and Y.~Yang, ``Network pruning via transformable architecture search,''
  in \emph{NeurIPS}, 2019, pp. 759--770.

\bibitem{luo2017thinet}
J.-H. Luo, J.~Wu, and W.~Lin, ``Thinet: A filter level pruning method for deep
  neural network compression,'' in \emph{ICCV}, 2017, pp. 5058--5066.

\bibitem{yu2018nisp}
R.~Yu, A.~Li, C.-F. Chen, J.-H. Lai, V.~I. Morariu, X.~Han, M.~Gao, C.-Y. Lin,
  and L.~S. Davis, ``Nisp: Pruning networks using neuron importance score
  propagation,'' in \emph{CVPR}, 2018, pp. 9194--9203.

\bibitem{liu2018rethinking}
Z.~Liu, M.~Sun, T.~Zhou, G.~Huang, and T.~Darrell, ``Rethinking the value of
  network pruning,'' in \emph{ICLR}, 2019.

\bibitem{wen2016learning}
W.~Wen, C.~Wu, Y.~Wang, Y.~Chen, and H.~Li, ``Learning structured sparsity in
  deep neural networks,'' in \emph{NeurIPS}, 2016, pp. 2074--2082.

\bibitem{he2017channel}
Y.~He, X.~Zhang, and J.~Sun, ``Channel pruning for accelerating very deep
  neural networks,'' in \emph{ICCV}, 2017, pp. 1389--1397.

\bibitem{zhuang2018discrimination}
Z.~Zhuang, M.~Tan, B.~Zhuang, J.~Liu, Y.~Guo, Q.~Wu, J.~Huang, and J.~Zhu,
  ``Discrimination-aware channel pruning for deep neural networks,'' in
  \emph{NeurIPS}, 2018, pp. 875--886.

\bibitem{he2018soft}
Y.~He, G.~Kang, X.~Dong, Y.~Fu, and Y.~Yang, ``Soft filter pruning for
  accelerating deep convolutional neural networks,'' in \emph{IJCAI}, 2018.

\bibitem{zhang2018systematic}
T.~Zhang, S.~Ye, K.~Zhang, J.~Tang, W.~Wen, M.~Fardad, and Y.~Wang, ``A
  systematic dnn weight pruning framework using alternating direction method of
  multipliers,'' in \emph{ECCV}, 2018, pp. 184--199.

\bibitem{li2019compressing}
T.~Li, B.~Wu, Y.~Yang, Y.~Fan, Y.~Zhang, and W.~Liu, ``Compressing
  convolutional neural networks via factorized convolutional filters,'' in
  \emph{CVPR}, 2019, pp. 3977--3986.

\bibitem{he2016deep}
K.~He, X.~Zhang, S.~Ren, and J.~Sun, ``Deep residual learning for image
  recognition,'' in \emph{CVPR}, 2016, pp. 770--778.

\bibitem{boyd2011distributed}
S.~Boyd, N.~Parikh, E.~Chu, B.~Peleato, J.~Eckstein \emph{et~al.},
  ``Distributed optimization and statistical learning via the alternating
  direction method of multipliers,'' \emph{Foundations and
  Trends{\textregistered} in Machine learning}, vol.~3, no.~1, pp. 1--122,
  2011.

\bibitem{ren2019admm}
A.~Ren, T.~Zhang, S.~Ye, J.~Li, W.~Xu, X.~Qian, X.~Lin, and Y.~Wang, ``Admm-nn:
  An algorithm-hardware co-design framework of dnns using alternating direction
  methods of multipliers,'' in \emph{ASPLOS}, 2019, pp. 925--938.

\bibitem{lecun1998mnist}
Y.~LeCun, C.~Cortes, and C.~J. Burges, ``The mnist database of handwritten
  digits, 1998,'' \emph{URL http://yann. lecun. com/exdb/mnist}, vol.~10,
  p.~34, 1998.

\bibitem{goodfellow2013empirical}
I.~J. Goodfellow, M.~Mirza, D.~Xiao, A.~Courville, and Y.~Bengio, ``An
  empirical investigation of catastrophic forgetting in gradient-based neural
  networks,'' \emph{arXiv preprint arXiv:1312.6211}, 2013.

\bibitem{krizhevsky2009learning}
A.~Krizhevsky, G.~Hinton \emph{et~al.}, ``Learning multiple layers of features
  from tiny images,'' 2009.

\bibitem{Jian-iot2020}
T.~Jian, B.~C. Rendon, E.~Ojuba, N.~Soltani, Z.~Wang, K.~Sankhe, A.~Gritsenko,
  J.~Dy, K.~Chowdhury, and S.~Ioannidis, ``Deep learning for rf fingerprinting:
  A massive experimental study,'' in \emph{IEEE Internet of Things Magazine},
  2020.

\bibitem{gritsenko2019finding}
A.~Gritsenko, Z.~Wang, T.~Jian, J.~Dy, K.~Chowdhury, and S.~Ioannidis,
  ``Finding a ‘new’needle in the haystack: Unseen radio detection in large
  populations using deep learning,'' in \emph{DySPAN}.\hskip 1em plus 0.5em
  minus 0.4em\relax IEEE, 2019, pp. 1--10.

\bibitem{goodfellow2014generative}
I.~Goodfellow, J.~Pouget-Abadie, M.~Mirza, B.~Xu, D.~Warde-Farley, S.~Ozair,
  A.~Courville, and Y.~Bengio, ``Generative adversarial nets,'' in
  \emph{NeurIPS}, 2014, pp. 2672--2680.

\bibitem{resnet}
K.~He, X.~Zhang, S.~Ren, and J.~Sun, ``Deep residual learning for image
  recognition,'' in \emph{CVPR}, 2016, pp. 770--778.

\bibitem{adam}
D.~P. Kingma and J.~Ba, ``Adam: A method for stochastic optimization,'' in
  \emph{ICLR}, 2015.

\bibitem{pytorch}
A.~Paszke \emph{et~al.}, ``Pytorch: An imperative style, high-performance deep
  learning library,'' in \emph{NeurIPS}, 2019, pp. 8024--8035.

\end{thebibliography}
\vspace{12pt}
\begin{appendices}
  \section{Solving Problem \eqref{eq:obj} via ADMM}\label{sec:admm-appendix}

To begin with, constraints \eqref{cons:non-overlap}, \eqref{cons:mask} are easy to satisfy: we basically partition variables of $W$ and $M$ to sets $\operatorname{supp}(\bar{W}^{t-1})$ and its complement, and only optimize over the appropriate set (the complement of $\operatorname{supp}(\bar{W}^{t-1})$ for $W$ and $\operatorname{supp}(\bar{W}^{t-1})$ for $M$). We thus ignore these constraints below. We similarly omit $W_{L+1}$, which is unconstrained and can be learned via SGD.
Rewriting the loss as $\mathcal{L}(W,M)$, we convert the non-convex optimization problem formulated in \eqref{eq:obj} into the ADMM form by introducing auxiliary variables $Z_l$ and $Y_l$ for constraints \eqref{cons:prune} and \eqref{cons:mask} respectively:%
\begin{subequations}%
\label{eq:admm-obj}%
\begin{align}%
    \underset{W,M}{\text{min:}}& \quad \textstyle\mathcal{L}(W,M)+\sum_{l=1}^L g_l(Z_l)+\sum_{l=1}^L h_l(Y_l),\displaybreak[0]\\
    \text{subject to:}& \quad W_l = Z_l,\quad l = 1, \cdots, L, \displaybreak[0]\\
    & \quad M_l = Y_l, \quad l = 1, \cdots, L,%
\end{align}%
\end{subequations}%
where $g_l(\cdot)$ and $h_l(\cdot)$ correspond to the indicator functions for constraints (\ref{cons:prune}) and (\ref{cons:mask}) respectively, i.e.,:
\begin{align}
    g_l(Z_l) = \begin{cases}
    0, \text{ if }\!{Z}_l\!\in\! S_l^t,\\
    +\infty,\text{ o.w.},
    \end{cases}\!
    h_l(Y_l) = \begin{cases} 
    0,\text{ if }\!{Y}_l\!\in\! S_l^{'t}, \\
    +\infty,\text{ o.w.}
    \end{cases}
\end{align}
%
The augmented Lagrangian of (\ref{eq:admm-obj}) is:
\begin{equation}\label{eq:admm-process}
    \begin{split}
        &\mathcal{L}_a(W, M, Z, U, Y, K) = \mathcal{L}(W,M) 
        \\
      &\!+\!\! \textstyle\sum_{l=1}^{L}\!\big\{ g_l(Z_l) \!+\! \rho_l \Tr({U}_l^\top (W_l\!-\!{Z}_l ) )\!+\! \frac{\rho_{l}}{2}   \|W_l\!-\!{{Z}}_{l} \|_{F}^{2} \big\} \\
      &\!+\!\! \textstyle\sum_{l=1}^{L}\! \big \{ h_l(Y_l) \! +\! \tau_l \Tr({K}_l^\top (M_l\!-\!Y_l) ) \!+\! \frac{\tau_{l}}{2}   \|M_l\!-\!Y_l \|_{F}^{2} \big\} \\
    \end{split}
\end{equation}
where ${\rho_{l}}$ and ${\tau_{l}}$ are penalty terms, and ${{U}}_{l}\in\reals^{P_l \times Q_l}$ and ${{K}}_{l}\in\reals^{P_l \times Q_l}$ are dual variables, rescaled by $\rho_l$ and ${\tau_{l}}$, respectively.
ADMM  proceeds iteratively as follows; at the $n$-th iteration: 
\begin{subequations}\label{eq:admm-process-ci}
    \begin{align}
        (W,M)^{n+1} 
      &\!=\! \underset{{W},{M}}{\argmin}\, \mathcal{L}_a({W},\! {M},\! {Z}^{n}\!\!,\!{U}^{n}\!\!,\! {Y}^{n}\!\!,{K}^{n})\label{eq:primal}\\
        ({Z}, {Y})^{n+1}&\!=\! \underset{{Z}, Y}{\argmin}\,\mathcal{L}_a({W^{n+\!1}}\!\!\!,\! {M^{n+1}}\!\!\!,\! {Z},{U}^{n}\!\!,\! {Y}\!,\!{K}^{n}\!)  \label{eq:proximal}\\
    {U}^{n+1} &= {U}^{n} + {W}^{n+1}-{Z}^{n+1} \\
    {K}^{n+1} &= {K}^{n} + {M}^{n+1}-{Y}^{n+1}.
    \end{align}
\end{subequations}

The problem \eqref{eq:primal} is equivalent to:  
\begin{equation} \label{eq:primal1}
    \begin{split}
    \min_{\bm{W},\bm{b}}~~ & \mathcal{L}(W,M)  
        + \textstyle\sum_{l=1}^L \frac{\rho_l}{2}\| W_l \!-\! Z_{l}^{n}\!+\!U_{l}^{n}\|_F^2 \\
        &+ \textstyle\sum_{l=1}^L \frac{\rho_l}{2}\| M_l\! -\! Y_{l}^{n}\!+\!K_{l}^{n}\|_F^2 .
    \end{split}
\end{equation}
The first term in  \eqref{eq:primal1} is a standard DNN loss while the second and the third terms are quadratic and differentiable. Thus, this subproblem can be solved by classic stochastic gradient descent.  
%
Problem \eqref{eq:proximal} is equivalent to:
%
\begin{subequations}
    \begin{align}\label{eq:admm-process-z}
    Z_l^{n+1} &= \textstyle\proj_{S_{l}^t}\big(W_l^{n+1}+U_{l}^{n}\big),~\text{ for all}~l=1,\ldots,L,\\
    \label{eq:admm-process-y}
    Y_l^{n+1} &= \textstyle\proj_{S_{l}^{'t}}\big(M_l^{n+1}+K_{l}^{n}\big),~\text{ for all}~l=1,\ldots,L,
    \end{align}
\end{subequations}
 where $\proj_{S_{l}^t}, \proj_{S_{l}^{'t}}$ are the Euclidean projections onto  sets $S_{l}, S_{l}^{'}$, respectively.

\section{Proof the correctness of Mask Projector}\label{sec:project-appendix}
For simplicity, we prove this for the projection to the set:
$S = \big\{x\in\{0,1\}^n: \|x\|_0 =k\}.$
i.e., the set of $n$ binary elements containing k zeros.
Let $y\in\mathbb{R}^n$, then $\proj_{S} (y)$ is computed by: 
(a) sort all elements $y_i \in y$ from smallest to largest; 
(b) set the $k$ largest values to 1 an the rest to 0.
We make use of the following lemma.
\begin{lemma}
    For $a,b \in \mathbb{R}$, where $a \leqslant b$,
    $a^2 + (1-b)^2 \leqslant (a-1)^2 + b^2.$
\end{lemma}
This can be easily proved by considering all positional cases of $a,b\in \reals$. Let $\hat{y}\in S$ be the solution of the algorithm, and $y^{*}\in\argmin_{x\in S} \|x-y\|_2$ be an optimal solution. Assume indices are order based on the elements of $y$, as in the algorithm. Let $i$ be the first position at which $\hat{y}_i\neq y^*_i$. Then, $y_i$ is mapped to 0 in $\hat{y}$ and $y_i$ is mapped to 1 in $y^{*}_i$. Moreover, as both have exactly $k$ ones, there must be a $j$ such that (i) $y_j \geqslant y_i$, (ii) $\hat{y}_j=1$, and (iii) $y_j^*=0$. By the lemma, since $y_j \geqslant y_i$, we have $y_i^2 + (y_j-1)^2 \leqslant (y_i-1)^2 + y_j^2$. So, setting $y^*_i=0$ and $y^*_j=1$ would only improve distance from $y$. As $y^*$ is optimal, this swap must maintain optimality; repeating this procedure as long as there exist indices at which $\hat{y}$ and $y^*$ differ will convert  $y^*$ to $\hat{y}$, while maintaining optimality. \qed 
\end{appendices}

\end{document}